\documentclass[sigconf]{acmart}

\usepackage{xspace}
\usepackage{fancyvrb} 
\usepackage{enumitem} 		   
\usepackage{tabularx,booktabs} 
\usepackage{bm}				   
\usepackage{mathtools} 		   
\usepackage{color,soul}        
\usepackage{amsmath}           
\usepackage{natbib}            
\usepackage{tikz}              
\usepackage{esvect}  
\usepackage{pifont}

\makeatletter
\DeclareRobustCommand*\cal{\@fontswitch\relax\mathcal}
\makeatother

\DeclareMathOperator*{\argmax}{argmax}

\newcolumntype{L}{>{\raggedright\let\newline\\\arraybackslash\hspace{0pt}}m{0.45\linewidth}}

\setcopyright{none}
\renewcommand\footnotetextcopyrightpermission[1]{} 
\begin{document}

\title{Identifying the Adoption or Rejection of Misinformation Targeting COVID-19 Vaccines in Twitter Discourse}


\author{Maxwell Weinzierl}
\orcid{0000-0002-8049-7453}
\affiliation{%
  \institution{Human Language Technology Research Institute, University of Texas at Dallas}
  \city{Richardson}
  \state{Texas}
  \country{USA}
}
\email{maxwell.weinzierl@utdallas.edu}

\author{Sanda Harabagiu}
\orcid{0000-0002-8186-1501}
\affiliation{%
  \institution{Human Language Technology Research Institute, University of Texas at Dallas}
  \city{Richardson}
  \state{Texas}
  \country{USA}
}
\email{sanda@utdallas.edu}

\renewcommand{\shortauthors}{Weinzierl and Harabagiu}

\begin{abstract}
Although billions of COVID-19 vaccines have been administered, too many people remain hesitant. Misinformation about the COVID-19 vaccines, propagating on social media, is believed to drive hesitancy towards vaccination. However, exposure to misinformation does not necessarily indicate misinformation adoption. In this paper we describe a novel framework for identifying the stance towards misinformation, relying on {\em attitude consistency} and its properties. The interactions between attitude consistency, adoption or rejection of misinformation and the content of microblogs are exploited in a novel neural architecture, where the stance towards misinformation is organized in a knowledge graph. This new neural framework is enabling the identification of stance towards misinformation about COVID-19 vaccines with state-of-the-art results. The experiments are performed on a new dataset of misinformation towards COVID-19 vaccines, called {\sc CoVaxLies}, collected from recent Twitter discourse. Because {\sc CoVaxLies} provides a taxonomy of the misinformation about COVID-19 vaccines, we are able to show which type of misinformation is mostly adopted and which is mostly rejected. 
 
\end{abstract}




\maketitle

\section{Introduction}

Although billions of inoculations against the SARS-CoV-2 virus, the causative agent of COVID-19, have been administered around the world starting with 2020, too many remain hesitant about this vaccine. 
It is believed that hesitancy is driven by misinformation about the COVID-19 vaccines that is spread on social media.
Recent research by \citet{nature-misinfo} has shown that exposure to online misinformation around 
COVID-19 vaccines affects intent to vaccinate in order to protect oneself and others. However, exposure
to misinformation about the COVID-19 vaccines does not mean that those exposed adopt the misinformation. This is why knowing if misinformation is adopted or rejected when encountered in social media discourse will enable public health experts to perform interventions at the right time and in the right place on social media, addressing vaccine hesitancy successfully. 

Misinformation detection on social media platforms, such as Twitter, is performed in two steps: (1) the  recognition whether a social media posting contains any misconception, reference to conspiracy theories
or faulted reasoning; and (2) the recognition of the {\em stance} 
towards the targeted misinformation. The stance defines the attitude the author of the micro-blog manifests towards the misinformation target, as exemplified in Table~\ref{tb:stance}. When the misinformation is adopted, an {\em Accept} stance is observed, whereas when it is rejected, the {\em Reject} stance reflects the attitude towards the targeted misinformation. 

    

\begin{table}[ht]
\centering
\small
\begin{tabular}{p{0.45\textwidth}}
    \toprule 
    {\bf Misinformation Target:} {\em The COVID vaccine renders pregnancies risky. }\\
    \midrule
    {\sc Stance:} {\bf Accept}\\
    {\em Tweet:} 
    <@USER> Chances of a healthy young woman dying of COVID if they even catch it: 0.003\% Chances of COVID vaccine causing miscarriage, birth defects, or future infertility: <Data Unavailable> Risk management would say DON'T TAKE THE VACCINE IF YOU'RE PREGNANT.
    \\
    \hline
    {\sc Stance:} {\bf Reject} \\
    {\em Tweet:} 
    Vaccinated women who breastfeed can pass \#COVID19 protection to their babies. COVID-19 \#vaccines aren’t considered a risk to infants during pregnancy or from breastfeeding. During the study, none of the women or infants experienced serious adverse events. <URL>
    \\ 
    \bottomrule
\end{tabular}
\caption{Examples of tweets with different stance towards misinformation targeting COVID-19 vaccines.}
\label{tb:stance}
\vspace{-6mm}
\end{table}
\raggedbottom
Although the identification of misinformation about COVID-19 vaccines in the Twitter discourse is fundamental in understanding its impact on vaccine hesitancy, we consider that efforts focusing on this first step of misinformation detection have made important progress recently, 
generating high-quality results \cite{Liu-aaai18, ma-etal-2017-detect, ma-etal-2018-rumor, Ma-GAN}.
In this paper we focus on the second step of misinformation detection, namely the identification of the stance towards misinformation, which still needs improvements. 

A significant barrier in the identification of stance towards misinformation targeting the COVID-19 vaccines stems from the absence of large Twitter datasets which cover misinformation about these vaccines. 
To address this limitation, we present in this paper a new Twitter dataset, called {\sc CoVaxLies}, inspired by the recently released {\sc COVIDLies} dataset \cite{covidlies}.  {\sc CoVaxLies} consists of (1) multiple known Misinformation Targets (MisTs) towards COVID-19 vaccines; (2) a large set of [tweet, MisT] pairs, indicating  when the tweet has the stance of: (a) {\em Accept} towards the MisT; (b) {\em Reject} towards the MisT; or (c)  {\em No Stance} towards the MisT. In addition, we provide a taxonomy of the misinformation about COVID-19 vaccines, informed by the MisTs available in {\sc CoVaxLies}, enabling the interpretation of the adopted or rejected misinformation about COVID-19 vaccines. 

As it can be noticed from the examples listed in Table~\ref{tb:stance}, 
identifying the stance of a tweet with respect to a given MisT is not a trivial language processing
task. 
The
 framework for stance identification presented in this paper makes several contributions that address
 the Twitter discourse referring to misinformation.
 First, it takes into account the {\em attitude consistency} (AC) observed throughout the Twitter discourse between tweet authors that adopt or reject a MisT. AC is informing the equivalence between stance identification and the recognition of {\em agree} or {\em disagree} relations between pairs of tweets. 
Second, this stance identification framework captures the interactions between discourse AC, the stance values of tweets towards a MisT, and the language used in the articulation of the MisT and the content of the tweets. 
Third, it considers that the Twitter discourse about a MisT encapsulates knowledge that can be represented by learning knowledge embeddings. This knowledge contributes, along with the neural representation of the content language 
of  tweets, to the prediction of agreement or  disagreement between pairs of tweets referring to the same MisT.
Finally, the system implementing this novel stance identification framework has produced in our experiments very promising results on the 
{\sc CoVaxLies} dataset. 


The remainder of the paper is organized as follows. Section~\ref{sec:related-work} describes the related work while Section~\ref{sec:covaxlies} details the {\sc CoVaxLies} dataset. Section~\ref{sec:methods} describes stance identification informed by attitude consistency (AC). Section~\ref{sec:results} presents the experimental result while Section~\ref{sec:dis} is providing discussions of the results. Section~\ref{sec:conclusion} summarizes the conclusions.


\section{Related Work}
\label{sec:related-work}

In previous work stance identification on Twitter was cast either as (1) a classification problem, learning to predict the stance value of a tweet towards 
a given target claim; or (2) an inference problem, concluding that a tweet entails, contradicts or does not imply the given target claim.

{\bf Stance identification as a classification problem:} Several datasets were used in prior work aiming stance classification on Twitter. 
The PHEME dataset \cite{Zubiaga2016} consists of Twitter conversation threads associated with 9 different newsworthy events such as the Ferguson unrest,  the shooting
at Charlie Hebdo, or Michael Essien contracting Ebola. A conversation thread consists of a tweet making a true and false claim, and a series of replies. There are 6,425 conversation threads in PHEME, 1,067 were annotated as true,  638 were annotated as false and 697 as unverified. A fraction of the PHEME dataset was used in the RumourEval task \cite{rumoureval}.
The stance labels are ‘support’, ‘deny’, ‘comment’ and ‘query’. There are 865 tweets annotated with the ‘support’ stance  label; 325 tweets annotated with the ‘deny’ stance label; 341 tweets annotated with the ‘query’ stance label and 2789 tweets annotated with the ‘comment’ stance label. 
Several neural classification architectures for stance identification were designed by participants in 
RumourEval \cite{augenstein-etal-2016-usfd, liu-etal-2016-iucl, wei-etal-2016-pkudblab}. 
However, \citet{Ghosh2019StanceDI} have shown that the
original pre-trained BERT \cite{bert} without any further fine-tuning outperforms all these former state-of-the-art
models on the RumourEval dataset, including the model that
utilizes both text and user information \cite{del-tredici-etal-2019-shall}. 

More recently, another dataset containing stance annotations was released, namely the {\sc COVIDLies} dataset \cite{covidlies}. The starting point was provided by 86 common misconceptions about COVID-19 available from the Wikipedia 
page dedicated to COVID-19 misinformation, which became Misinformation Targets (MisTs). 
For each known MisT, a set of tweets were 
annotated with three possible values for stance towards each misconception: (1) agree, when the
tweet adopts the MisT; (2) disagree, when the tweet contradicts/rejects the MisT; and (3) no stance when the tweet is either neutral or is irrelevant
to the MisT. Of the 6761 annotated tweets, 5,748
(85.02\%) received a label of no stance; 670 (9.91\%) received
a label of agree and 343 (5.07\%) received a label of disagree. Recently, using this dataset, \citet{our-stance}
used a neural language processing model that exploits the pre-trained domain-specific language
model COVID-Twitter-BERT-v2 \cite{Muller2020COVIDTwitterBERTAN} and refined it by stacking several layers of lexico-syntactic, semantic, and emotion Graph Attention Networks (GATs) \cite{velikovi2017graph}  to learn and all the possible interactions between these different linguistic phenomena, before classifying a tweet as (a) agreeing; (b) disagreeing or (c) having no stance towards a MisT.

{\bf Stance identification as an inference problem:}
When the {\sc COVIDLies} dataset of stance annotations was released in \cite{covidlies}, stance identification
was presented as a natural language inference problem which can benefit from existing textual inference datasets.
In fact, Bidirectional LSTM encoders and Sentence-BERT (SBERT) \cite{sentence-bert} were trained on three common NLI datasets—SNLI \cite{bowman-etal-2015-large}, MultiNLI \cite{mnli}, and MedNLI \cite{mednli}. 

We were intrigued and inspired by the {\sc COVIDLies} dataset, and believed that we could create a similar dataset containing misinformation about COVID-19 vaccines, which would not only complement the {\sc COVIDLies} data,
but it would also enable the development of novel techniques for identifying the stance towards misinformation targeting COVID-19 vaccines.  

 \begin{figure*}[t]
    \centering
    \includegraphics[width=0.9\textwidth]{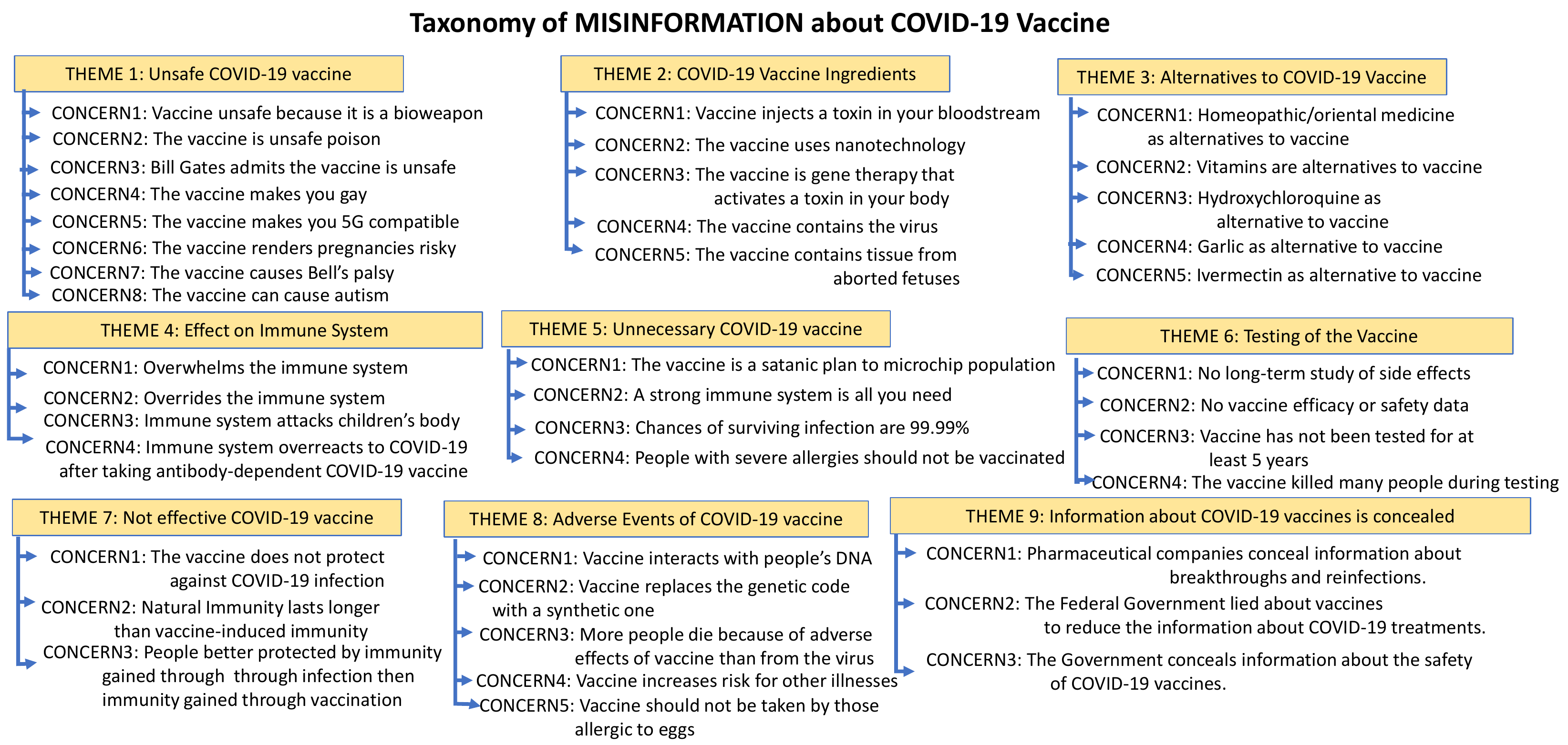}
    \caption{Taxonomy of Misinformation}
    \label{fig:taxonomy}
\vspace{-4mm}
\end{figure*}
 
\section{Stance Annotations in CoVaxLies} \label{sec:covaxlies}
\subsection{CoVaxLies: A Twitter Dataset of Misinformation about COVID-19 Vaccines} \label{sec:data}

The {\sc CoVaxLies} Twitter dataset contains misinformation about COVID-19 vaccines represented as (1) several {\em known} Misinformation Targets (MisTs); (2) a collection of tweets paired with the MisTs they refer to and annotated with stance values, indicating whether the tweet agrees, disagrees or has no stance towards the MisT; and (3) a taxonomy of misinformation about the COVID-19 vaccines, revealing the 
themes and the concerns addressed by the MisTs from {\sc CoVaxLies}.
We used two information sources for identifying Misinformation Targets (Mists) for COVID-19 vaccines.
First, we have considered (a) the Wikipedia page available at {\em en.wikipedia.org/wiki/COVID-19\_misinformation\#Vaccines}, which collects many misconception claims referring to the vaccines developed for immunization against the SARS-CoV-2 virus; and (b) MisTs identified by  
organizations such as the Mayo Clinic, University of Missouri Health Care, University of California (UC) Davis Health, University of Alabama at Birmingham, Science-Based Medicine, PublicHealth.org, Snopes, and the British Broadcast Corporation (BBC), which have been actively collecting misinformation about the COVID-19 vaccines and debunking them on public websites. 
There are 17 MisTs about COVID-19 vaccines identified in this way in {\sc CoVaxLies}. Appendix~\ref{ax:mists} provides examples of MisTs identified in this way.

Secondly, we have used 19 questions from the Vaccine Confidence Repository \cite{Rossen} to retrieve 
answers from an index of 5,865,046 unique original tweets obtained from the Twitter streaming API as a result of the query “(covid OR coronavirus) vaccine lang:en”. These tweets were authored in the time frame from December 18th, 2019, to January 4th, 2021. Many answers that were retrieved as responding to questions about vaccine confidence contained misinformation, and became MisTs as well. 
We identified an additional set of 37 MisTs, out of which 7 MisTs were already known to us from the first source of information. Appendix~\ref{ax:mists} provides examples of MisTs retrieved as answers to questions about vaccine confidence.
Therefore, {\sc CoVaxLies} relies on  47 MisTs about COVID-19 vaccines. 
Before using the Twitter streaming API to collect tweets discussing the COVID-19 vaccine, approval from the Institutional Review Board at the University of Texas at Dallas was obtained: IRB-21-515 stipulated that our research met the criteria for exemption.

In order to identify $\cal{T}_R$, the collection of  tweets which potentially contain language relevant to the MisTs from {\sc CoVaxLies}, we relied on two information retrieval systems: (1) a retrieval system using the BM25 \cite{bm25} scoring function; and (2) a retrieval system using  BERTScore \cite{bertscore} with Domain Adaptation (DA), identical to the one used by \citet{covidlies}. 
Both these retrieval systems operated on an index of $\cal{C}_{T}$,  retrieving tweets by processing   {\sc CoVaxLies} MisTs as queries.

Researchers from the Human Language Technology Research Institute (HLTRI) at the University of Texas at Dallas judged 7,346 tweets to be relevant to the  MisTs from
{\sc CoVaxLies} and organized them in [tweet, MisT] {\em pairs}, annotated with stance information.
There are 3,720 tweets which {\em Accept} their MisT, 2,194 tweets which {\em Reject} their and 1,238 tweets that have  {\em No Stance}. We note that {\sc CoVaxLies}
contains an order of magnitude more stance annotations than PHEME \cite{Zubiaga2016}, the most popular Twitter dataset containing stance annotations, and therefore it presents clear advantages for neural learning methods.

To enable the usage of {\sc CoVaxLies} in neural learning frameworks, we split the tweets into three distinct collections: (a) a training collection; (b) a development collection; and (c) a test collection. The training collection, which consists of 5,267 [tweet, MisT] pairs, was utilized to train our automatic stance identification systems, described in Section~\ref{sec:methods}. The development collection, which consists of 527 [tweet, MisT] pairs, was used to select model hyperparameters, such as threshold values.
The test collection, which consists of 1,452 [tweet, MisT] pairs, was used to evaluate the stance identification approaches, enabling us to report the results in Section~\ref{sec:results}.


\begin{figure}[t]
    \centering
    \includegraphics[width=0.8\linewidth]{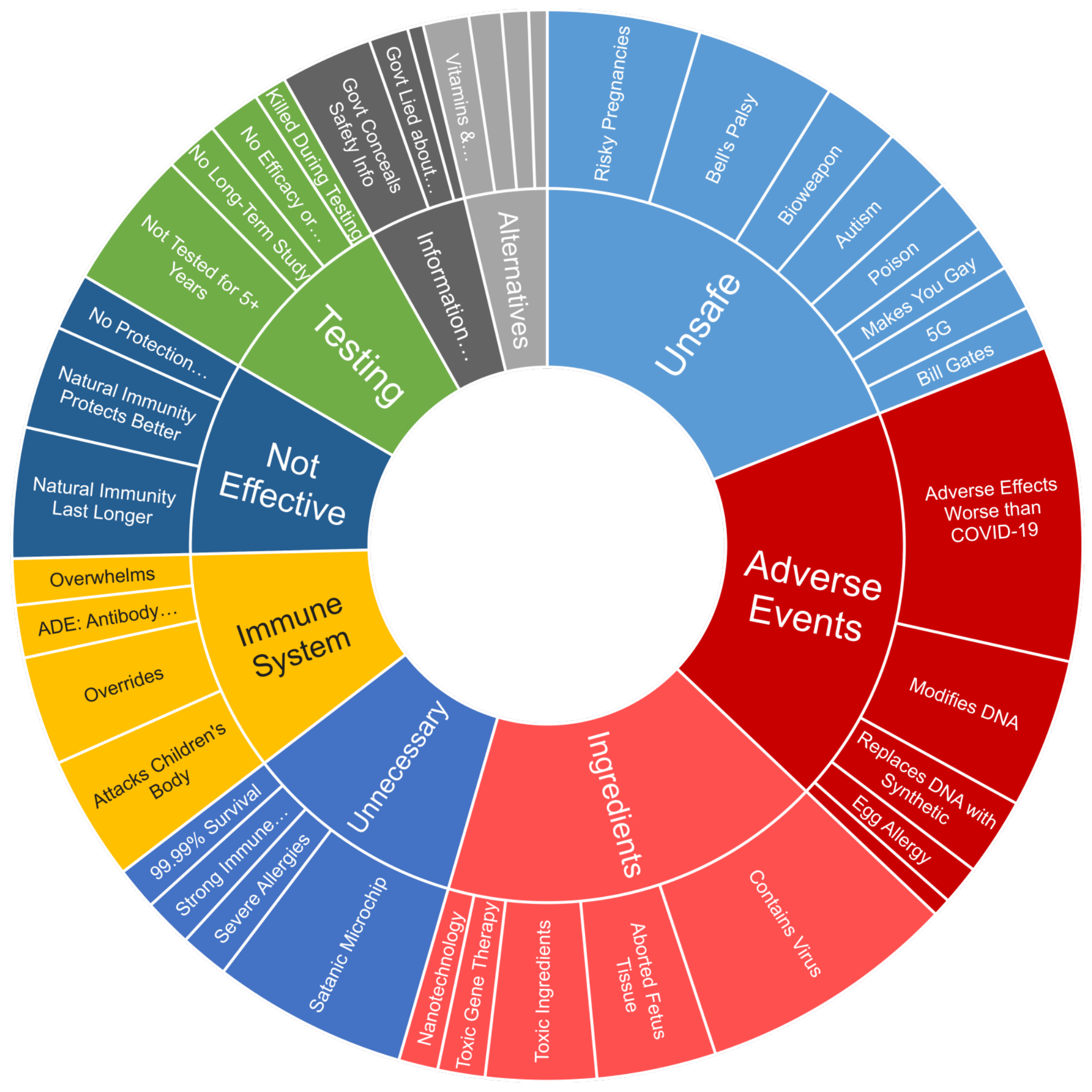}
    \caption{Distribution of  Misinformation Themes and Concerns in the tweets available from \sc{CoVaxLies}.}
    \label{fig:theme_concern_size}
    
\vspace{-4mm}
\end{figure}

\subsection{The Misinformation Taxonomy from CoVaxLies}
\label{sec:taxonomy}
Figure~\ref{fig:taxonomy} illustrates the taxonomy of misinformation available in {\sc CoVaxLies}. The themes represent the highest level of abstraction, while the concerns differentiate the various MisTs from {\sc CoVaxLies}. 
The taxonomy emerged from discussion between public health experts from the University of California, Irvine School of Medicine and computational linguists from HLTRI. Nine misinformation themes were revealed, all characterizing aspects that impact confidence in the COVID-19 vaccine.
Confidence, along with convenience and complacency, are well known universal factors contributing to vaccine hesitancy, according to the 3C model \cite{3C}. For each misinformation theme, as shown in Figure~\ref{fig:taxonomy}, a different number of concerns were revealed: the largest number of concerns pertain to the theme predicating the fact that the COVID-19 vaccines are unsafe (8 concerns) while the smallest number of concerns pertain to the themes claiming that the vaccines are not effective or that information about the vaccines is concealed. 
Using the information provided by the taxonomy
illustrated in Figure~\ref{fig:taxonomy}, we notice in Figure~\ref{fig:theme_concern_size} that 
the misinformation themes that dominate the tweets from {\sc CoVaxLies} are those about the ingredients of the COVID-19 vaccines, about the adverse events and the fact that the vaccines are unsafe. Moreover, the dominant misinformation regarding the vaccine ingredients claims that the vaccines contain the virus,
while the dominant concerns of the lack of safety of the vaccines indicates risky pregnancies or Bell's palsy.

\begin{figure}[t]
    \centering
    \includegraphics[width=0.8\linewidth]{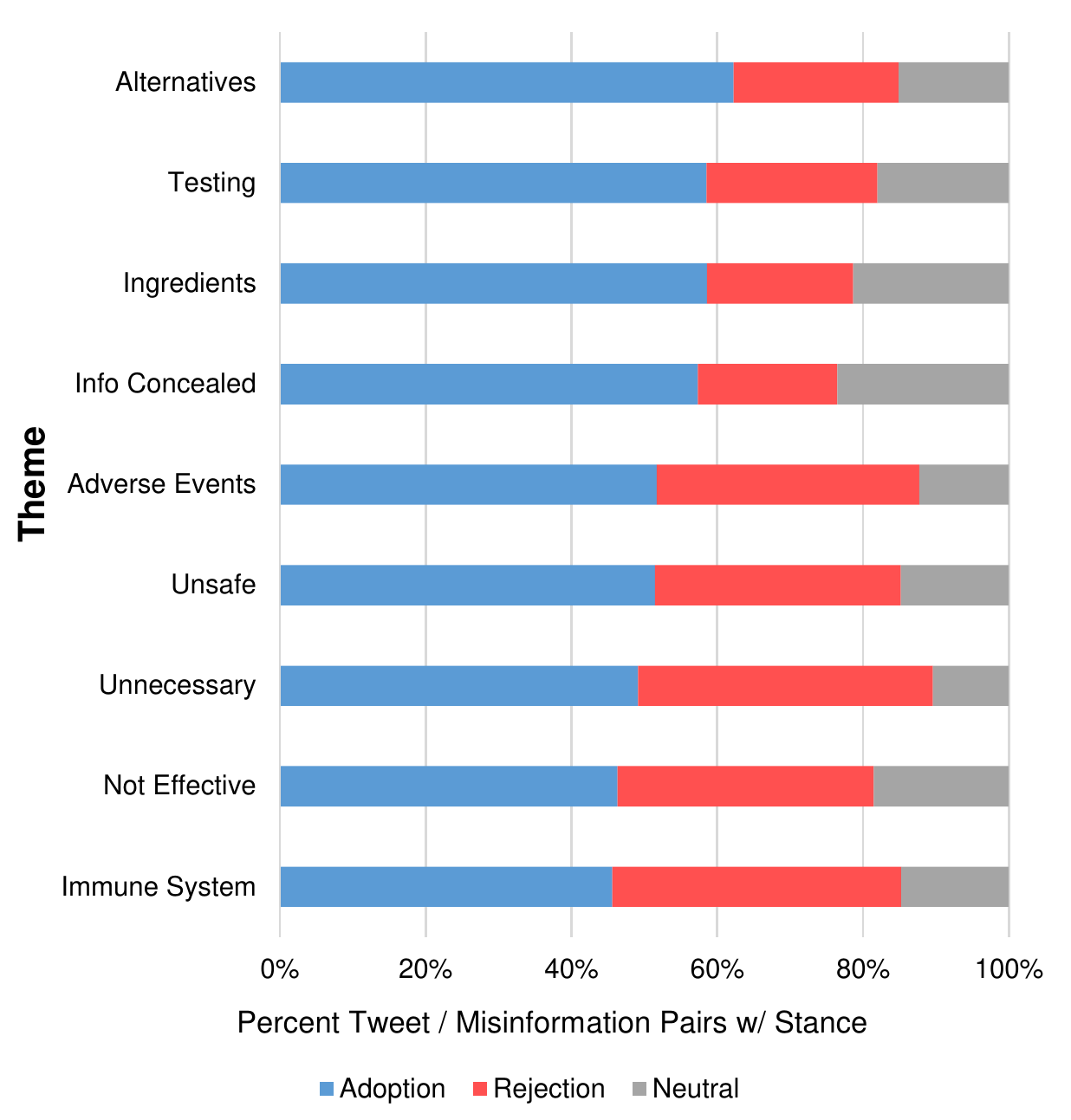}
    \caption{Distribution of Stance Values across Misinformation Themes in {\sc CoVaxLies}. }
    \label{fig:theme_stance}
    
\vspace{-4mm}
\end{figure}

When considering the distribution of tweets that adopt the misinformation, those that reject it and those that are neutral (because of having no stance) for the tweets across all the misinformation themes, we noticed, as illustrated in Figure~\ref{fig:theme_stance}, that the misinformation that is most adopted has the theme of considering alternatives to the COVID-19 vaccines, immediately followed by misinformation regarding the testing of the vaccines and the ingredients used in the vaccines. Interestingly, most of the misinformation that is rejected has to do with the theme indicating that the COVID-19 vaccines are unnecessary, or that they affect the immune system.


\section{Stance Identification through Attitude Consistency}
\label{sec:methods}
\subsection{Attitude Consistency and Stance}
\label{sec:ac}

\begin{figure}[ht]
\centering
\includegraphics[width=0.30\textwidth]{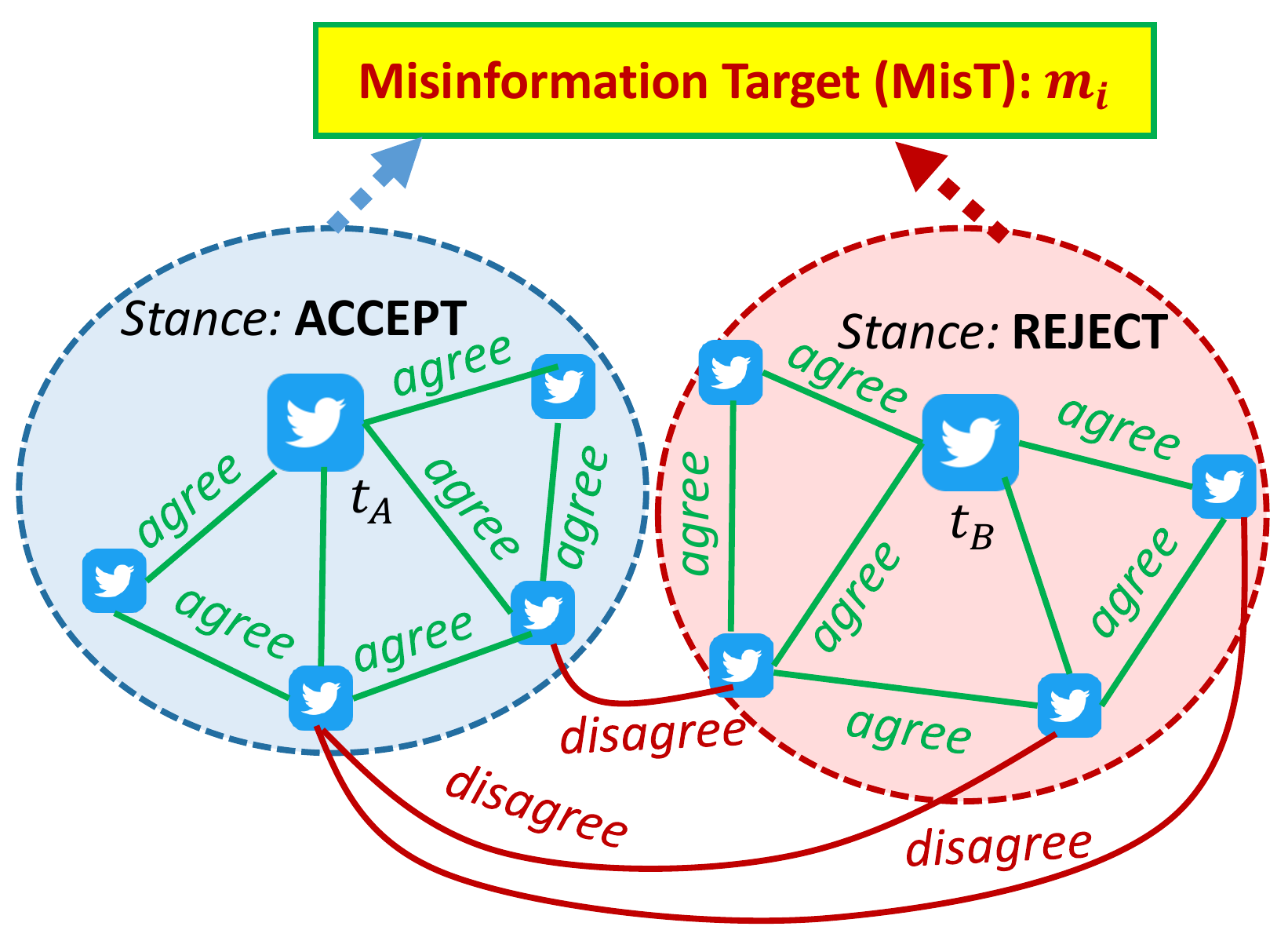}
\caption{Stance Misinformation Knowledge Graph}
\label{fig:smkg}
\vspace{-4mm}
\end{figure}

Central to our stance identification framework is the belief that 
the stance of any tweet $t_j$ towards a particular MisT $m_i$ should not be considered in isolation. 
Because $t_j$ participates in the Twitter discourse about $m_i$, its stance should be consistent with the attitude of the other tweet authors towards $m_i$. We hypothesize that all the authors of tweets that {\em Accept} $m_i$ must be agreeing among themselves with regard to $m_i$. Similarly, all the authors of tweets that {\em Reject} $m_i$ must also be agreeing among themselves with regard to $m_i$.  But also, any author of a tweet $t_j$ that has an {\em Accept} stance towards $m_i$ must disagree with the author of any tweet $t_k$ that has a {\em Reject} stance towards $m_i$. Therefore, all these tweet authors have Attitude Consistency (AC) towards $m_i$. AC can be illustrated as in Figure~\ref{fig:smkg},  by linking
all the tweets that have the {\em same stance} towards a MisT $m_i$ through implicit {\em agree} relations, and all tweets that have {\em opposing stances} towards $m_i$ with implicit {\em disagree} relations. 
In this way, all the tweets that have an {\em Accept} stance towards $m_i$ are organized in a fully connected graph spanned by {\em agree} relations and similarly, all the tweets having a {\em Reject} stance towards $m_i$ are organized in a fully connected graph spanned also by {\em agree} relations. In addition, {\em disagree} relations are established between all pairs of tweets that have opposing stance towards $m_i$. Moreover, all
tweets that do not have either an {\em Accept} or {\em Reject} stance towards $m_i$ are considered to automatically have {\em No Stance} towards $m_i$. Hence. the stance values $SV=\{Accept, Reject\}$ are the only ones informing AC.

\begin{figure}[t]
\centering
\includegraphics[width=0.35\textwidth]{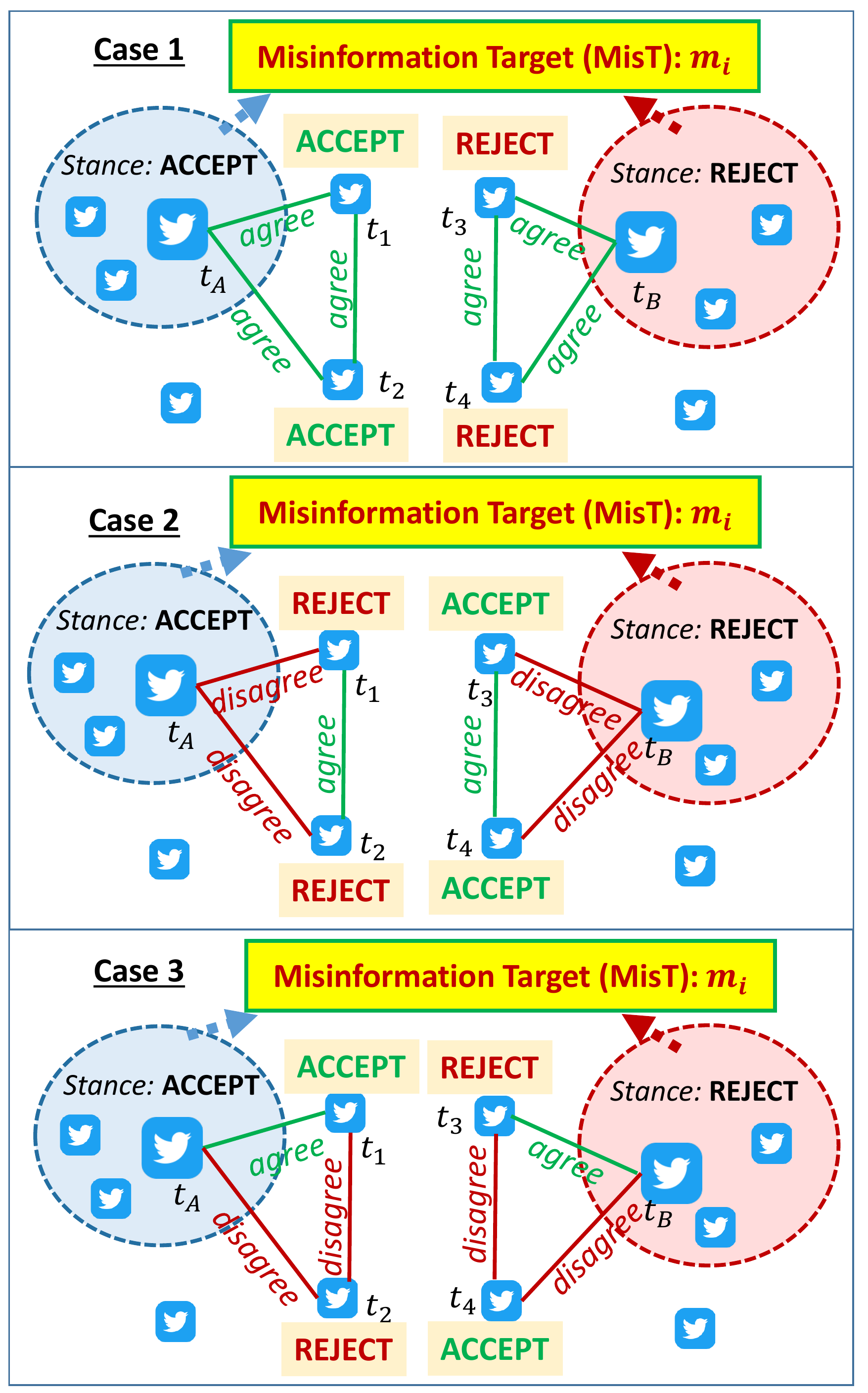}
\caption{Attitude Consistency Examples}
\label{fig:c_examples}
\vspace{-4mm}
\end{figure}

As shown in Figure~\ref{fig:smkg}, a Stance Misinformation Knowledge Graph is organized for each $m_i$, referred to as SMKG($m_i$). For clarity, the
SMKG($m_i$) illustrated in Figure~\ref{fig:smkg} shows only several of the {\em agree} and {\em disagree} relations. 
For each MisT $m_i$ available in the {\sc CoVaxLies} dataset, we generate an SMKG($m_i$) 
when considering only the tweets annotated with {\em Accept} or {\em Reject} stance information, available from the training set of {\sc CoVaxLies}.  
However, there are many other tweets in {\sc CoVaxLies} with no known stance towards any of the MisTs available in the dataset. We refer to the entire set of such tweets as the Tweets with Unknown Stance towards Misinformation (TUSM). 

To identify the stance of tweets from TUSM we assume that AC is preserved. This entails three possible cases when considering in addition to the SMKG($m_i$),  tweets from TUSM, e.g. $t_1$, $t_2$, $t_3$ or $t_4$, as illustrated in  Figure~\ref{fig:c_examples}. 
All the three cases of AC show that the unknown stance of any tweet $t_x \in$ TUSM can be identified as {\em Accept}  when knowing if (a) an {\em agree} relation is predicted between $t_x$ and $t_A$, a tweet known to have an {\em Accept} stance towards $m_i$; 
or (b) a {\em disagree} relation is predicted between $t_x$ and $t_B$, a tweet known to have a {\em Reject} stance towards $m_i$. Similarly, the unknown stance of any tweet $t_x \in$ TUSM can be identified as {\em Reject}  when knowing if (a) a {\em disagree} relation is predicted between $t_x$ and $t_A$,  a tweet known to have an {\em Accept} stance towards $m_i$; or (b)  an {\em agree} relation is predicted between $t_x$ and $t_B$, a tweet known to have a {\em Reject} stance towards $m_i$.  If none of these relations can be predicted, then the stance of $t_x$ is
identified as {\em No Stance} towards $m_i$. To formalize the interaction between the implicit relation types 
and the values of the stance towards a MisT $m_i$ identified for a pair of tweets $t_x$ and $t_y$ we
considered a function that
selects the Relation Type that preserves AC (RTAC), defined as: 
\begin{equation} \label{eq:rtac}
    RTAC(s_x, s_y)= 
    \left \{ \begin{array}{ll}
agree & \mbox{if $s_x = s_y$} \\
disagree & \mbox{if $s_x \neq s_y$} 
\end{array}
\right. 
\end{equation}
where the value of the stance of $t_x$ towards $m_i$ is $s_x$ while the value of the stance of $t_y$ is $s_y$.
Moreover, we believe that AC can be further extended to account for an entire chain of {\em agree} and {\em disagree} relations spanning tweets with unknown stance towards $m_i$.

\subsection{Transitive Attitude Consistency}
\label{sec:acs}
\begin{figure}[ht]
\centering
\includegraphics[width=0.46\textwidth]{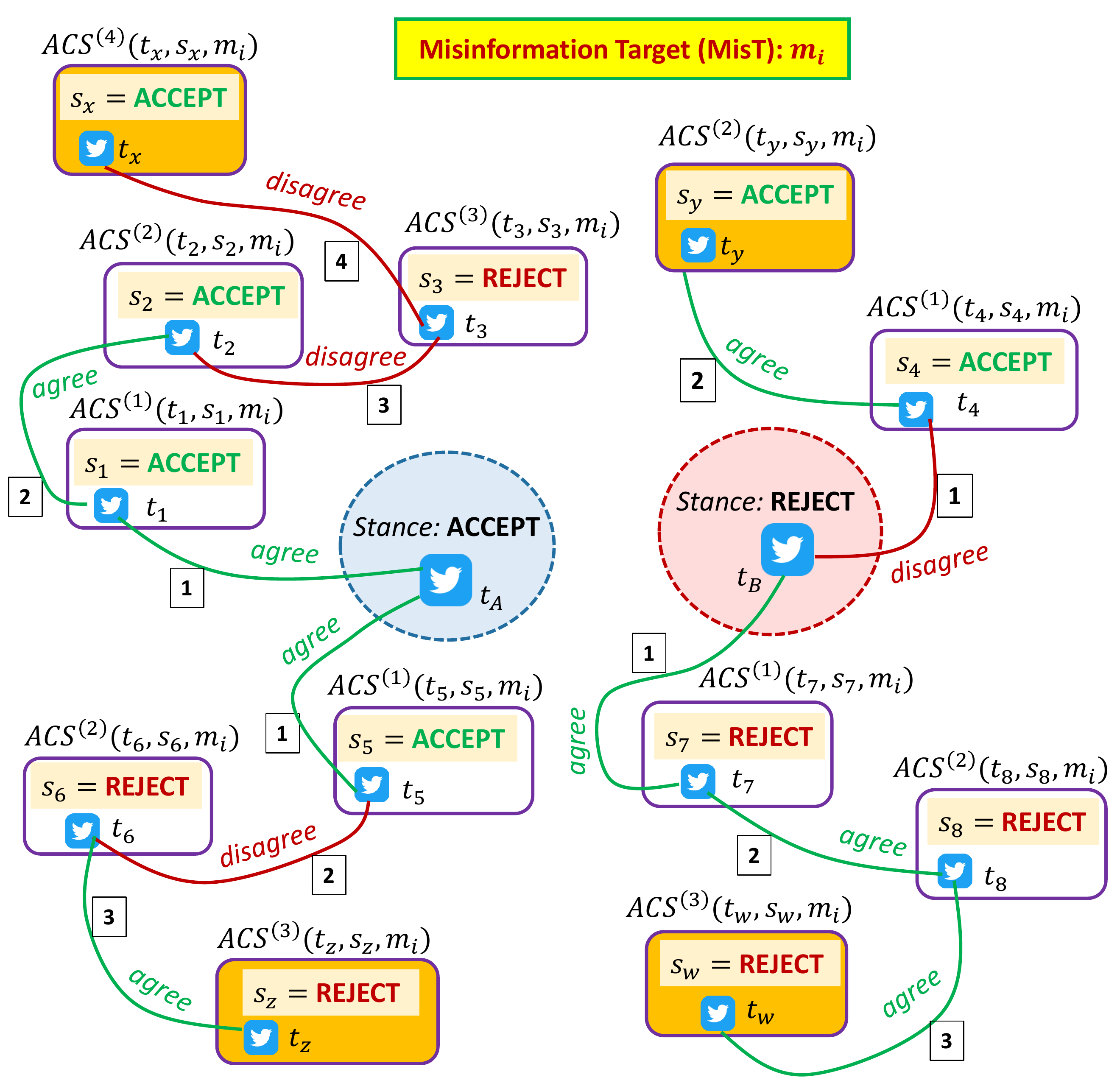}
\caption{Stance Identification with Transitive Attitude Consistency and Attitude Consistency Scores.}
\label{fig:steps}

\end{figure}

Transitive Attitude Consistency extends the interaction between the values of the stance towards a MisT $m_i$ and
the binary {\em agree} and {\em disagree} relations to an entire chain of such implicit relations that may connect a tweet from TUSM  to a tweet from SMKG($m_i$), whose stance is known. For example, Figure~\ref{fig:steps} shows how the identified stance towards $m_i$ of tweets $t_x$, $t_y$, 
$t_z$ and $t_w$ is informed by chains of {\em agree} or {\em disagree} relations originating
either in $t_A$ or $t_B$, tweets from SMKG($m_i$). It is important to note that this 
extension has to take into account that every time a new stance $s_x$ towards a MisT $m_i$ is identified for a tweet $t_x \in$ TUSM, the confidence that the AC is preserved is computed by an Attitude Consistency Score ($ACS$).
$ACS$ depends on $l$, the number of relations in the chain originating at a tweet with known stance, available from SMKG($m_i$) and ending 
at a tweet $t_x \in$ TUSM, with unknown stance: $ACS^{l}(t_x, s_x, m_i)$. To compute $ACS^{l}(t_x, s_x, m_i)$ we first need to consider the way in which we can represent the SMKG($m_i$).

The knowledge graph of SMKG($m_i$) can be represented in a continuous vector space called the {\em embedding space} by learning {\em knowledge embeddings} for its nodes and edges. When formalizing the 
SMKG($m_i$)$=(V; E)$, 
each node $v_k \in V$ can be seen as $v_k=(t_k,s_k)$, where a tweet $t_k$ is paired with its stance $s_k$ towards $m_i$; and each edge $e_{ij} \in E$ is either an {\em agree} or a {\em disagree} relation.
Knowledge embedding models learn an embedding $te_k$ for each tweet $t_k$ as well as an embedding $me_i^{agree}$
for the {\em agree} relation in SMKG($m_i$)  and an embedding $me_i^{disagree}$
for the {\em disagree} relation in SMKG($m_i$). But more importantly, knowledge embedding models use a relation scoring
function $f$ for assigning a plausibility score to any potential link
between two tweets $t_x$ and $t_y$, given their knowledge embeddings $te_x$ and $te_y$ and the embedding of the relation they share. Because the relation between $t_x$ and $t_y$ must preserve AC between the stance $s_x$ identified for $t_x$ and the stance $s_y$ identified for $t_y$, the relation between these two tweets is provided by the function $RTAC(s_x, s_y)$. 
The embedding of the relation indicated by $RTAC(s_x, s_y)$ is computed as:
\begin{equation} \label{eq:re}
    RE(s_x, s_y,m_i)= 
    \left \{ \begin{array}{ll}
me_i^{agree} & \mbox{if $RTAC(s_x, s_y)$ = {\em agree}} \\
me_i^{disagree} & \mbox{if $RTAC(s_x, s_y)$ = {\em disagree}} 
\end{array}
\right. 
\end{equation}
Hence, the scoring function of the relation between the pair of tweets $t_x$ and $t_y$ is defined as $f(te_x,RE(s_x,s_y,m_i),te_y)$, where $f$ is provided by various knowledge embedding models, such as those 
that we discuss in Section~\ref{sec:ke-models},

Given the representation of SMKG($m_i$) through knowledge embeddings, we can define $ACS^{l}(t_x, s_x, m_i)$, starting with the chains of length $l=1$: 
\begin{equation}
    ACS^{1}(t_x, s_x, m_i)= 
    \sum\limits_{(t_y, s_y) \in SMKG(m_i)}
       \frac{f(te_x, RE(s_x, s_y, m_i), te_y)}{|SMKG(m_i)|}
\end{equation}
Then, $ACS^{l}(t_x, s_x, m_i)$ for chains of length $l>1$ is computed by considering that we have defined already $SV=\{Accept, Reject\}$ and that we shall take into account all tweets from TUSM when generating chains of {\em agree} and/or {\em disagree} relations originating in SMKG. 
We compute $ACS^{l}(t_x, s_x, m_i)$ as:
\begin{multline}
    ACS^{l}(t_x, s_x, m_i)= \\
    \sum\limits_{\substack{
        t_z\in TUSM\\ 
        t_z \neq t_x}
    }
        \sum\limits_{s_z \in SV}
            \frac{ACS^{l-1}(t_z, s_z, m_i) +
            f(te_x, RE(s_x, s_z, m_i), te_z)}{|TUSM|-1}
\end{multline}
To consider the overall $ACS^*$ 
of any tweet $t_x$ with stance $s_x$ towards $m_i$ we average the $ACS$  across all possible chains of relations, of varying lengths, up to a maximum length $L$:
\begin{equation} \label{eq:acs_star}
    ACS^*(t_x, s_x, m_i)=
    \frac{1}{L}
    \sum\limits_{l=1}^{L}
        ACS^l(t_x, s_x,m_i)
\end{equation}
Finally, stance $s_x$ towards $m_i$ of a tweet $t_x \in$ TUSM is assigned the value corresponding to the maximum $ACS^*$:
\begin{equation}
    s_x = \argmax_{s_k \in SV}{ACS^*(t_x, s_k,m_i)}
\end{equation}
However, Equation~\ref{eq:acs_star} shows how we assign stance of value {\em Accept} or {\em Reject} to tweets with previously unknown stance towards a MisT $m_i$. To also assign the stance value {\em No Stance}, we relied on the development set from {\sc CoVaxLies} to assign a threshold value $T(m_i)$ for each MisT $m_i$,
such that when $ACS^*(t_x, s_x, m_i) \leq T(m_i)$, for stance values {\em Accept} and {\em Reject}, we can finalize the stance $s_x$ of a tweet $t_x$ as having the value {\em No Stance}.  With all stance values finalized for tweets from TUSM towards any
MisT $m_i$ from {\sc CoVaxLies}, we update SMKG($m_i$) to contain all the tweets from TUSM that have either an {\em Accept} or a {\em Reject} stance towards $m_i$.



\subsection{Learning Knowledge Embeddings for the Stance Misinformation Knowledge Graph}
\label{sec:ke-models}

Knowledge embedding models such as TransE~\cite{transe} and TransD~\cite{transd} have had significant success in modeling relations in knowledge graphs. 
More recently, new knowledge embeddings models capture more complex interactions from the knowledge graph, e.g.  TransMS~\cite{transms}, TuckER~\cite{tucker}, and RotatE~\cite{rotate}.
Each knowledge embedding model provides a different method of scoring the likelihood of relations in the knowledge graph SMKG($m_i$), as shown in Table~\ref{tb:ke}.
The scoring of a relation in each knowledge embedding model relies on $me_i^r$, 
the embedding of a relation that maintains AC with the 
stance towards a MisT $m_i$ of the tweets connected by the relation, and on the embeddings of these tweets, $te_x$ and $te_y$.

\begin{table}[ht]
\centering
\small
\begin{tabular}{|l|c|}
    \toprule
    KE Model &  Scoring Function $f(te_x, me_i^r, te_y)$  \\
    \midrule
    {\bf TransE}~\cite{transe} &
    $
    -||te_x + me_i^r - te_y|| 
    $
    \\
    \hline
    {\bf TransD}~\cite{transd} &
    $
    -\left\lVert (\bm{I} + me_i^{r,p} \times ({te^p_x})^{\top}) \times te_x\right. + me_i^r
    $
    \\
    &
    $
    - \left. (\bm{I} + me_i^{r,p} \times ({te^p_y})^{\top} ) \times te_y \right\rVert 
    $
    \\
    \hline
    {\bf TransMS}~\cite{transms} &
    $
    -\left\lVert - tanh(te_y \odot me_i^r) \odot te_x 
    \right. + me_i^r
    $
    \\
    &
    $
    + \alpha_i^r \cdot (te_x \odot te_y) 
    \left. 
    - tanh(te_x \odot me_i^r) \odot te_y \right\rVert 
    $
    \\
    \hline
    {\bf TuckER}~\cite{tucker} &
    $
    \mathcal{W} \times_1 te_x \times_2 me_i^r \times_3 te_y
    $
    \\
    \hline
    {\bf RotatE}~\cite{rotate} &
    $
    -||te_x \odot me_i^r - te_y|| 
    $
    \\
    \bottomrule
\end{tabular}
\caption{Knowledge Embedding Scoring Functions.}
\label{tb:ke}
\vspace{-6mm}
\end{table}
\raggedbottom

In Table~\ref{tb:ke}, we denote  
$||\cdot||$ as the $L1$ norm, $\bm{I}$ is the identity matrix, $tanh(x)$ is the non-linear hyperbolic tangent function and 
$\alpha^r_i$ is a real numbered parameter dependent on each MisT. 
The operator $\odot$ represents the Hadamard product, and $\times_n$ indicates the tensor product along the n-th mode.
Any of the scoring functions listed in Table~\ref{tb:ke} measure the likelihood of an {\em agree} or
{\em disagree} relation between a pair of tweets which preserves the AC with the stance of the tweets. However, the
content of the tweets, communicated in natural language, with the subtleties and deep connections expressed in language also need to be captured when scoring these relations.

\subsection{Interactions between Tweet Language, Stance towards Misinformation and Attitude Consistency}
\label{sec:language}

\begin{figure}[ht]
    \centering
    \includegraphics[width=0.47\textwidth]{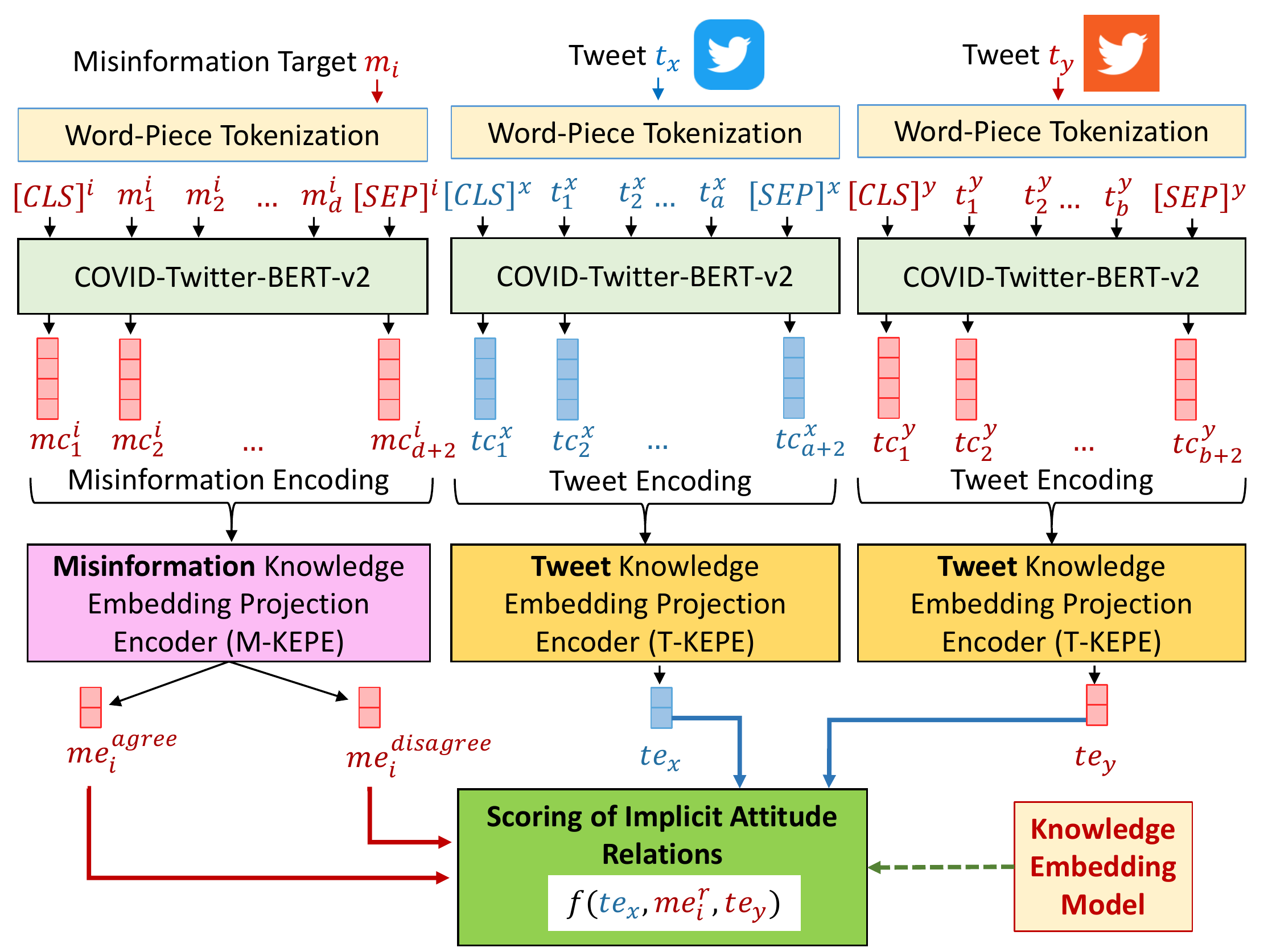}
    \caption{Neural Architecture for Language-informed Attitude Consistency-preserving Relation scoring (LACRscore).}
    \label{fig:architecture}
    
\end{figure}

The AC of various tweet authors is expressed through the language they use in their tweets. Therefore. it is imperative to also consider the interaction of the language of tweets with the stance towards misinformation and the attitude consistency of the tweet author's discourse. Because the identification of stance towards misinformation is equivalent to discovering the type of relation, either {\em agree} or {\em disagree}, shared by a pair of tweets that preserves AC, we designed a neural language architecture 
which considers (1) the contextual embeddings of each MisT $m_i$, as well as each pair of tweets $t_x$ and $t_y$ having a stance towards $m_i$; and (2) knowledge embeddings learned for the SMKG($m_i$) such that we predict the likelihood of a relation between 
$t_x$ and $t_y$ to be of type {\em agree} or to be of type {\em disagree}. This neural architecture for Language-informed Attitude Consistency-preserving Relation scoring (LACRscore) is illustrated in Figure~\ref{fig:architecture}. 

Given a MisT $m_i$, the LACRscore system first performs Word-Piece Tokenization \cite{bert} on (a) the textual description of $m_i$, producing tokens $m_1^i, m_2^i, ..., m_d^i$, as well as on the text of tweets $t_x$ and $t_y$, 
which are then passed through the BERT \cite{bert} COVID-19 Language Model COVID-Twitter-BERT-v2 \cite{covid-twitter-bert}
pre-trained on the masked language modeling task \cite{bert} for 97 million COVID-19 tweets. 
This process of further pre-training has been shown to improve performance on downstream tasks in various scientific \cite{scibert}, biomedical \cite{biobert}, and social media \cite{bertweet} domains. 
COVID-Twitter-BERT-v2 produces contextualized embeddings $mc_1^i, mc_2^i, ..., mc_{d+2}^i$ for the word-piece tokens in the MisT $m_i$ along with the $[CLS]^i$ and $[SEP]^i$ tokens.  
In this way, 
we encode the language describing the MisT $m_i$ using a contextualized embedding $mc_1^i \in \mathbb{R}^{1024}$, where $1024$ is the contextual embedding size for COVID-Twitter-BERT-V2.
Similarly, the language used in the tweets $t_x$ and $t_y$ is 
represented by contextual embeddings $tc_1^x$ and $tc_1^y$ after being processed through COVID-Twitter-BERT-v2.
But, it is important to note, that the scoring function $f$ from any of the knowledge embedding models 
provided in Table~\ref{tb:ke}, cannot operate directly on the contextual embeddings $tc_1^x$, $mc_1^i$ or $tc_1^y$, as they do not have the same dimensions of the knowledge embeddings these models learn. 
Additionally, we need to produce two knowledge embeddings for the MisT $m_i$ to represent both the {\em agree} and {\em disagree} relation embeddings.
Therefore, in LACRscore we needed to consider two forms of projection encoders, capable to project from the contextualized embedding space into the knowledge embedding space. 
For this purpose, we have relied on the Misinformation Knowledge Embedding Projection Encoder (M-KEPE), using two separate fully-connected layers, to project from $mc_1^i$ into the necessary knowledge embeddings $me_i^{agree}$ and $me_i^{disagree}$ from any of the knowledge embedding models considered. 
Similarly, the Tweet Knowledge Embedding Projection Encoder (T-KEPE) uses a different fully-connected layer than M-KEPE to project from $tc_1^x$ and $tc_1^y$ to $te_x$ and $te_y$ respectively. 
As shown in Figure~\ref{fig:architecture}, these encoders produce the arguments of the scoring function $f$, provided by some knowledge embedding model.
The likelihood of an {\em agree} or {\em disagree} relation between tweets $t_x$ and $t_y$ with respect to the MisT $m_i$ is computed by $f(te_x, me_i^{agree}, te_y)$ and $f(te_x, me_i^{disagree}, te_y)$. 

LACRscore was trained on the SMKG($m_i$) derived from the training collection of {\sc CoVaxLies}, described in Section~\ref{sec:data}. 
Relations from each SMKG($m_i$) were used as positive examples,  and we performed negative sampling to construct ``Attitude Inconsistent'' examples. 
Negative sampling consists of corrupting a relation $r$ between tweets $t_x$ with stance $s_x$ and $t_y$ with stance $s_y$ towards MisT $m_j$, which preserves AC. 
This corruption process is performed by randomly sampling either 
(1): a different tweet $(t_z, s_z) \in SMKG(m_i)$ with the same relation $\hat{r}=r$, to replace $t_y$ such that $RTAC(s_z, s_x) \neq r$, or 
(2): flipping $r$ from an {\em agree} relation to $\hat{r}=${\em disagree} relation, or vice versa. 
The negative sampling will ensure that AC relations will be scored higher than non-AC relations.
Moreover, we optimized the following margin loss to train LACRscore when scoring  relations:
\begin{equation} \label{eq:loss}
    \mathcal{L} =\\
    \sum\limits_{\substack{
    }}{
    \left[\gamma - f(te_x,me^r_i,te_y) + f(te_x,me_i^{\hat{r}},te_z)\right]_+}
\end{equation}
where $\gamma$ is a training score threshold which represents the differences between the score of AC relations and the non-AC relations. The loss $\mathcal{L}$ is minimized with the ADAM\cite{kingma2014adam} optimizer, a variant of gradient descent.

\begin{table*}[t]
\centering
\small
\begin{tabular}{lrrrrrrrrr}
\toprule
System & Accept F1 & Accept P & Accept R & Reject F1 & Reject P & Reject R & Macro F1 & Macro P & Macro R \\
\midrule
NLI-Tweet-MisT \cite{covidlies} & 45.9 & 72.9 & 33.5 & 54.6 & 38.6 & \textbf{93.2} & 50.2 & 55.8 & 63.3 \\
DS-StanceId  \cite{our-stance} & 86.2 & 88.3 & 84.2 & 79.1 & 82.7 & 75.8 & 82.7 & 85.5 & 80.0 \\
LES-GAT-StanceId  \cite{our-stance} & 86.7 & 84.6 & 88.9 & 80.7 & \textbf{83.2} & 78.3 & 83.7 & 83.9 & 83.6 \\
\hline
LACRscore & & & & & & & & & \\
  + TransE & 69.4 & 65.6 & 73.7 & 47.7 & 52.3 & 43.9 & 58.6 & 59.0 & 58.8 \\
  + TransE + ACS & 60.1 & 64.0 & 56.7 & 50.5 & 44.7 & 58.1 & 55.3 & 54.4 & 57.4 \\
  + TransD & 54.9 & 59.4 & 51.0 & 46.6 & 40.3 & 55.2 & 50.7 & 49.9 & 53.1 \\
  + TransD + ACS & 51.6 & 56.7 & 47.4 & 41.5 & 35.3 & 50.5 & 46.6 & 46.0 & 48.9 \\
  + TuckER & 87.7 & 86.7 & 88.7 & 82.3 & 79.3 & 85.5 & 85.0 & 82.0 & 87.1 \\
  + TuckER + ACS & 86.1 & 85.6 & 86.6 & 80.9 & 73.5 & 89.8 & 83.5 & 79.6 & 88.2 \\
  + RotatE  & 86.6 & 83.6 & 89.9 & 80.9 & 73.5 & 89.8 & 83.7 & 78.5 & 89.9 \\
  + RotatE + ACS & 86.6 & 85.7 & 87.5 & 83.0 & 80.5 & 85.8 & 84.8 & 83.1 & 86.6 \\
  + TransMS  & 85.7 & 81.8 & \textbf{90.0} & 78.4 & 69.3 & 90.3 & 82.1 & 75.6 & \textbf{90.1} \\
  + TransMS + ACS & \textbf{88.7} & \textbf{89.8} & 87.6 & \textbf{85.6} & \textbf{83.2} & 88.2 & \textbf{87.1} & \textbf{86.5} & 87.9 \\
\bottomrule
\end{tabular}
\caption{Results from the stance identification experiments on the {\sc CoVaxLies} dataset.}
\label{tb:results}
\vspace{-6mm}
\end{table*}
\raggedbottom

\section{Experimental Results}
\label{sec:results}
To evaluate the quality of stance identification on the test collection from {\sc CoVaxLies} we use the Precision (P), Recall (R), and F$_1$
metrics for detecting the \emph{Accept} and \emph{Reject} values of stance. We also compute a Macro averaged Precision, Recall, and F$_1$ score. The evaluation results 
are listed in Table~\ref{tb:results}.
The bolded numbers represent the best results obtained.
When evaluating the LACRscore system, we have considered (1) five possible knowledge embedding models (TransE; TransD; TuckER; RotatE; and TransMS), which provide different relation scoring functions; and (2) two possible options of stance prediction: (a) using the Attitude Consistency Scoring (ACS) approach described in Section~\ref{sec:acs}; and (b) ignoring ACS by and constraining $L=1$ for any chain of relations, thus ignoring the transitive property of AC.  

In addition, we have evaluated several baselines. 
First, we considered the system introduced by \citet{covidlies}, listed as the Natural Language Inference between Tweet text and MisT text (NLI-Tweet-MisT) system.
As a baseline, we have also considered the Domain-Specific Stance Identification (DS-StanceId) \cite{our-stance} system, which utilizes the "[CLS]" embedding from COVID-Twitter-BERT-v2 to directly perform stance classification. In addition, we considered  the Lexical, Emotion, and Semantic Graph Attention Network for Stance Identification (LES-GAT-StanceId) \cite{our-stance} system which
relies on Lexical, Emotion, and Semantic Graph Attention Networks.

The NLI-Tweet-MisT system produced a Macro F$_1$ score of $50.2$, indicating that stance identification as inference over language is not sufficient. 
Far superior results were obtained by the DS-StanceId system with a Macro F$_1$ score of $82.7$, showcasing the advantage of fine-tuning stance identification systems. 
The LES-GAT-StanceId system produced a Macro F$_1$ score of $83.7$, which indicates that integrating Lexical, Emotional, and Semantic Graphs further improves stance identification. 
The LACRscore system with the TuckER configuration produced a Macro F$_1$ score of $85.0$, indicating that identifying the stance towards misinformation through AC presents performance advantages over previous methods.  
Unsurprisingly. the LACRscore system with the TransMS + ACS configuration performed best, producing a Macro F$_1$ score of $87.1$, which indicates that the transitive nature of AC should not be ignored. 
The results also show that detecting misinformation rejection tends to be more difficult than the identification of misinformation adoption.

System hyperparameters were selected by maximizing the F$_1$ score of each system on the development set. 
The LACRscore system was trained with the following hyperparameters: a linearly decayed learning rate of $1e-4$ which was warmed up over the first $10\%$ of the $36$ total epochs, an attention drop-out rate of $10\%$, a batch size of $32$, and the tweet and MisT knowledge embedding size was set to $8$ for all knowledge embedding models, as we found that to perform best on the development set.
The LACRscore system utilized the training set for learning to score AC-preserving relations by optimizing the margin loss, described in Equation~\ref{eq:loss}.
The LACRscore system with the ACS configuration utilized a maximum chain length $L$ of $32$, the length value performing best on the development set. 
The $\gamma$ hyperparameter is set to $4.0$ for all knowledge graph embedding models, and we sampled $1$ negative corrupted relation for each AC relation in the SMKG($m_i$).
Threshold values $T(m_i)$ were also automatically selected by maximizing the F$_1$ score of the LACRscore system on each MisT $m_i$ on the development set.

\section{Discussion}
\label{sec:dis}

\begin{figure}[t]
    \centering
    \includegraphics[width=0.96\linewidth]{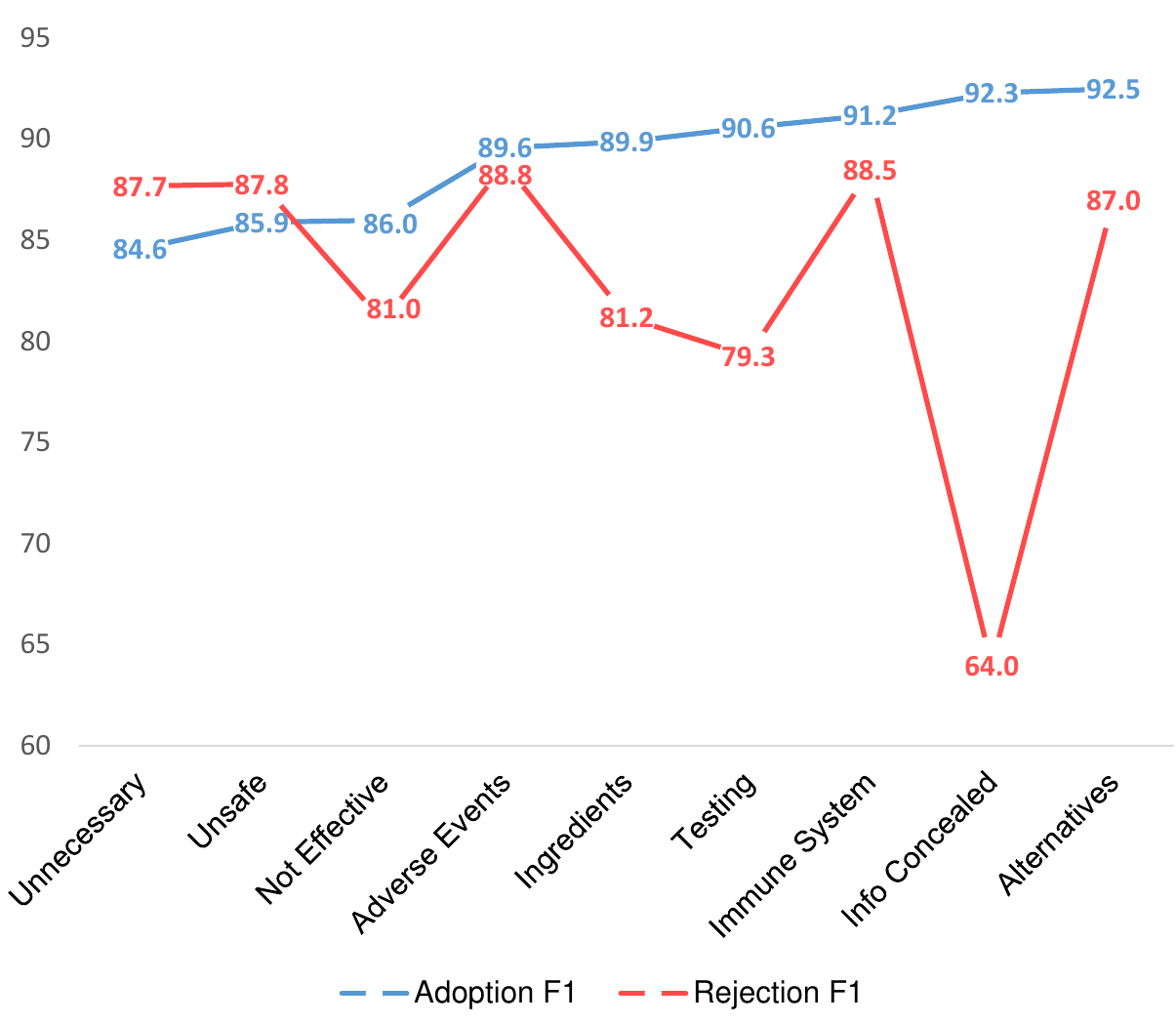}
    \caption{F$_1$-scores of the misinformation adoption vs. rejection discovered by the LACRscore system with the TransMS and ACS configuration across misinformation Themes from the {\sc CoVaxLies} dataset.}
    \vspace{-2mm}
    \label{fig:theme_performance}
\end{figure}

Because the LACRscore system produced the best results with the TransMS and ACS configuration,
we performed an analysis of the F$_1$ scores of this system across each of the themes 
available in the {\sc CoVaxLies} Misinformation Hierarchy, considering both the {\em adoption} and {\em rejection} of misinformation, as illustrated in Figure~\ref{fig:theme_performance}. The identification of adopted misinformation has remarkable performance, across all themes. Moreover, the misinformation rejection is identified quite well too, except for the theme of concealing information about vaccines. This is explained by the observation that this theme is addressed by few tweets in {\sc CoVaxLies}, as illustrated in Figure~\ref{fig:theme_concern_size}, and moreover, it has the smallest percentage of {\em rejection} stance values, as illustrated in Figure~\ref{fig:theme_stance}.

\section{Conclusion}
\label{sec:conclusion}
In this paper we present a new method for identifying the stance towards misinformation informed by attitude consistency (AC), which accounts for very promising results on {\sc CoVaxLies}, a new Twitter dataset of misinformation targeting the COVID-19 vaccines. AC proves to be a stronger signal for stance identification than lexical, emotional and semantic knowledge alone. Moreover, AC informs the knowledge encapsulated in the misinformation discourse on Twitter, which explains the promising results produced by this method, both for the adoption and rejection of misinformation about COVID-19 vaccines.
\bibliographystyle{ACM-Reference-Format}
\bibliography{web.bib}

\newpage
\appendix
\section{Misinformation Targets in CoVaxLies} 
\label{ax:mists}
Misinformation Targets (MisTs), which represent common misconceptions about the COVID-19 vaccines or refer to conspiracy theories associated with these vaccines, have two different sources.
In Table~\ref{tb:misinfo}, all examples
marked with $\diamond$ correspond to some of the MisTs identified as known misinformation from Wikipedia and other trusted sources, while all examples marked with $\Box$ correspond to some of the answers to questions about vaccine confidence, originating from \citet{Rossen}.

\begin{table}[h]
\centering
\small
\begin{tabular}{|p{0.95\linewidth}|}
    \toprule 
    $\diamond$ RNA alters a person's DNA when taking the COVID-19 vaccine.\\   \hline
    $\diamond$ The COVID-19 vaccine causes infertility or miscarriages in women.\\ \hline
    $\diamond$ The COVID-19 vaccine causes Bell's palsy.\\ \hline
    $\diamond$ The COVID-19 vaccine contains tissue from aborted fetuses. \\   \hline
    $\diamond$ The COVID-19 vaccine can cause autism. \\     \hline 
    $\diamond$ Hydroxychloroquine protects against COVID-19. \\ \hline
    $\diamond$The COVID-19 Vaccine is a satanic plan to microchip people\\  \hline
    $\Box$ There are severe side effects of the COVID-19 vaccines, worse than having the virus. \\ \hline
    $\Box$ The COVID-19 vaccine is not safe because it was rapidly developed and tested. \\     \hline
    $\Box$ The COVID-19 vaccine can increase risk for other illnesses.\\ \hline
    $\Box$ Vaccines contain unsafe toxins such as formaldehyde, mercury or aluminum.\\ \hline
    $\Box$ Governments hide COVID-19 vaccine safety information \\ \hline
    $\Box$ The COVID-19 Vaccine will make you gay. \\
    \bottomrule
\end{tabular}
\caption{Examples of COVID-19 {\sc Misinformation Targets}}
\label{tb:misinfo}
\end{table}

\section{Code and CoVaxLies Data Availability} 
\label{ax:code}
The {\sc CoVaxLies} dataset, comprising the Misinformation Targets (MisTs), the misinformation taxonomy, and the [${tweet}_i$, ${MisT}_j$] pairs, which associate a ${tweet}_i$ with its evoked ${MisT}_j$ along with stance annotations. 
The  {\sc CoVaxLies} dataset is publicly available at the following GitHub repository:

{\flushleft{\em \url{https://github.com/Supermaxman/vaccine-lies/tree/master/covid19}}}\\

{\flushleft Code needed to reproduce} the experiments described in this paper is also publicly available at the following GitHub repository:

{\flushleft{\em \url{https://github.com/Supermaxman/covid19-vaccine-nlp}}}\\

We note that an early version of {\sc CoVaxLies} was presented 
in \citet{weinzierl-covid19-glp}, but in that version of {\sc CoVaxLies} only 17 Misinformation Targets (MisTs) were available, namely the MisTs discovered from Wikipedia and other trusted sources, which are available in this later version as well. 
Moreover, the previous version of {\sc CoVaxLies} did not contain any \emph{stance} annotations, and it did not contain the misinformation taxonomy which were made available in the current version.

\end{document}